\documentclass[letterpaper]{article} 
\usepackage{aaai2027}  
\usepackage[hyphens]{url}  
\usepackage{graphicx} 
\usepackage{natbib}  
\usepackage{caption} 
\usepackage{algorithm}
\usepackage{algorithmic}
\usepackage{multirow}

\usepackage{booktabs} 
\usepackage{pifont}   
\usepackage{amsmath}
\usepackage[table]{xcolor}
\usepackage{tabularx}
\usepackage{amssymb}
\newcommand{\cmark}{\ding{51}} 
\newcommand{\xmark}{\ding{55}} 

\definecolor{reasoning_gray}{RGB}{128, 128, 128}
\definecolor{correct_green}{RGB}{34, 139, 34}
\definecolor{error_red}{RGB}{220, 20, 60}

\usepackage{newfloat}
\usepackage{listings}
\DeclareCaptionStyle{ruled}{labelfont=normalfont,labelsep=colon,strut=off} 
\floatstyle{ruled}
\newfloat{listing}{tb}{lst}{}
\floatname{listing}{Listing}

\usepackage{booktabs}
\usepackage{amsmath}
\usepackage{amssymb}

\title{DeceptionX: From Multimodal Evidence to Explainable Deception Detection}
\author{
    Jiayu Zhang\textsuperscript{\rm 1}\equalcontrib,
    Shuo Ye\textsuperscript{\rm 1}\equalcontrib,
    Shuo Ye\textsuperscript{\rm 1},
    Jiajian Huang\textsuperscript{\rm 1},
    Yawen Cui\textsuperscript{\rm 1},
    Taorui Wang\textsuperscript{\rm 1},
    Wei Xia\textsuperscript{\rm 1},
    Zeheng Wang\textsuperscript{\rm 1},
    Haowen Tang\textsuperscript{\rm 1},
    Yelin Wang\textsuperscript{\rm 1},
    Hui Ma\textsuperscript{\rm 1},
    Zitong Yu\textsuperscript{\rm 1}\corresponding\\
}
\affiliations{
    \textsuperscript{\rm 1}Great Bay University\\
    \textsuperscript{\rm 2}Hong Kong Polytechnic University\\
    
}

\begin{document}

\maketitle

\begin{abstract}
Deception detection is a challenging task within affective computing and behavioral analysis. Existing deep learning methods typically treat this task as a straightforward classification problem. However, traditional deep binary~(truthful/deceptive) models lack interpretability and fails to capture the complex logical deduction processes utilized by human experts when identifying lies. While Multimodal Large Language Models (MLLMs) show promise, applying them effectively requires a bridge between mid-level audiovisual cues and high-level logical reasoning. In this paper, we propose DeceptionX, a novel MLLM framework that shifts the paradigm of deception detection from black-box classification to an interpretable Observe-Think-Summarize reasoning process. To address the scarcity of high-quality reasoning data, we first constructed DeceptChain, a dataset developed through a human-in-the-loop process. This dataset synthesizes fine-grained visual and auditory evidence (such as micro-expressions and vocal tremors) into structured chain-of-thought reasoning data. Furthermore, we propose a three-stage training pipeline and a Discrepancy-Aware Redundancy Elimination~(DARE) strategy for DeceptionX to further enhance the model's generalization capabilities. Extensive experiments demonstrate that DeceptionX not only outperforms existing MLLM baselines and state-of-the-art methods on standard real-world benchmarks but also provides transparent, expert-level reasoning paths, bridging the critical gap between accuracy and interpretability in multimodal deception detection.
\end{abstract}


\begin{figure}[h]
  \centering
  \includegraphics[width=\linewidth]{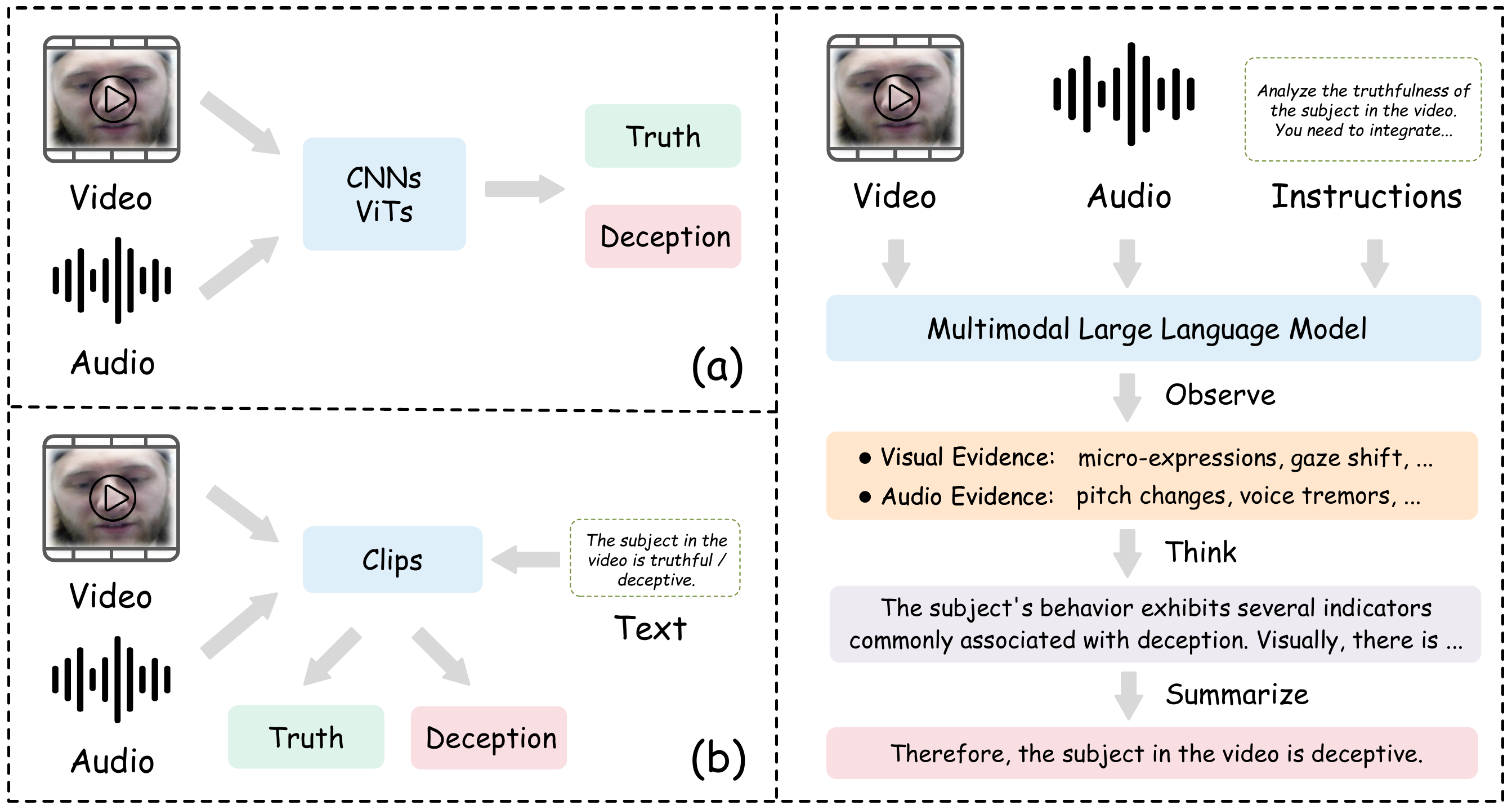}
  \vspace{-0.5cm}
  \caption{Compared to traditional feature-based~(a) and alignment-based~(b) paradigms, DeceptionX shifts from black-box binary classification to an explainable framework that mimics human cognition. By integrating multimodal cues including micro-expressions and voice tremors, DeceptionX follows an Observe-Think-Summarize pipeline to derive reasoned and transparent deception analysis.}
  \vspace{-0.5cm}
  \label{fig:1}
\end{figure}

\section{Introduction}
Deception~\cite{masip2017deception} refers to the act of intentionally misleading others. It is a common and complex human behavior that can pose significant threats to national, political, judicial, and economic security. For instance, in the context of public safety, criminals may conceal their intentions to evade detection, thereby endangering social security. Within the judicial system, the falsification of evidence or the provision of false statements can distort case outcomes, leading to wrongful convictions or acquittals, undermining legal authority, and eroding public trust. Therefore, the detection and prevention of deception are crucial for maintaining social stability and economic order. Because human judgment is often susceptible to cognitive biases~\cite{bond2006accuracy}, researchers are dedicated to developing multimodal automated deception detection systems. These systems identify potential deceptive behaviors by analyzing visual and auditory cues~\cite{perez2015deception}, including facial micro-expressions and vocal patterns.

Despite significant advancements in existing deep learning-based deception detection methods~\cite{wu2018deception,karnati2021lienet,nam2023facialcuenet,zhang2026multimodal}, they typically treat the task as a straightforward black-box binary classification problem, as illustrated in Figure~\ref{fig:1}. This paradigm suffers from two distinct limitations. First, black-box models lack interpretability, failing to capture the complex logical reasoning processes that human experts rely on when identifying lies. In high-stakes real-world applications, a standalone predictive label is far from sufficient; investigators require a verifiable chain of evidence that links low-level audio-visual cues~(e.g., gaze shifts, variations in vocal tone) to the final decision. Second, high-level visual cognitive tasks face the challenge of weak semantic correlation between discrete labels and visual content. Traditional classification labels struggle to encompass the subtle, dynamic nuances inherent in deceptive behavior.

In recent years, Multimodal Large Language Models~(MLLMs) have demonstrated significant potential for semantic understanding and reasoning~\cite{yang2025qwen3,achiam2023gpt}. However, existing MLLMs primarily focus on enhancing general capabilities such as video question answering, video captioning, object segmentation, and video content understanding. Their ability to discern deceptive behavior in videos remains limited. Consequently, effectively applying these models to the field of deception detection faces two major challenges. First, there is a lack of comprehensive and structured instructional data capable of clearly identifying, analyzing, and integrating potential deceptive behaviors within videos. Simply prompting off-the-shelf MLLMs often results in superficial judgments devoid of solid evidential grounding. Second, appropriate training strategies to distinctly enhance the model's analytical and deductive reasoning capabilities in this highly specialized domain have yet to be fully explored.

To address these challenges, we propose an automated construction method and introduce a new dataset named DeceptChain. To reduce the difficulty of annotation, we primarily rely on large language model generation with human assistance to create an automated annotation pipeline. The entire process is conducted in stages where we first extract detectable evidence of suspicious behavior from both visual and auditory perspectives, then further investigate the underlying deceptive processes based on this evidence, and finally merge all annotated attributes to form a coherent chain of reasoning. To bridge the gap between judgment accuracy and interpretable reasoning, we also propose a multimodal deception detection model named DeceptionX based on DeceptChain along with a progressive training strategy. Specifically, we first align audiovisual evidence with multimodal features within the text space, then employ instruction tuning to teach the model basic reasoning regarding the subjects' behavior in videos and response formatting, and finally utilize reinforcement learning to enhance its accuracy and reasoning rationality in deception detection. Furthermore, we introduce a Discrepancy-Aware Redundancy Elimination~(DARE) strategy to enhance the model's attention toward suspicious deceptive segments. In summary, our main contributions include:
\begin{itemize}
    \item We construct DeceptChain, the first high-quality deception detection dataset developed through a comprehensive human-in-the-loop process, which addresses the scarcity of reasoning data in deception detection by synthesizing fine-grained visual and auditory evidence into structured and logical chain-of-thought formats.
    \item We propose DeceptionX, a novel MLLM framework that shifts the deception detection paradigm from uninterpretable black-box classification to a transparent, human-like Observe-Think-Summarize reasoning process.
    \item We design a progressive three-stage training pipeline and introduce a Discrepancy-Aware Redundancy Elimination (DARE) strategy to significantly enhance the model's analytical focus and generalization capabilities.
    \item Extensive experiments demonstrate that DeceptionX outperforms existing MLLM baselines and state-of-the-art methods across multiple benchmarks, while providing transparent, expert-level reasoning paths.
\end{itemize}

\section{Related Works}

\subsection{Automated Deception Detection}

Automated deception detection~\cite{constancio2023deception,d2024analysis} represents a formidable and essential task within the fields of affective computing and behavioral analysis. The developmental trajectory of this field has transitioned from traditional feature engineering~\cite{hearst1998support, de2013decision, breiman2001random} to sophisticated deep learning architectures. While early research relied heavily on psychological priors and manually crafted statistical features, the widespread adoption of deep neural networks~\cite{abouelenien2018gender,bai2019automatic,monaro2022detecting,li2021survey,greff2016lstm,khan2022transformers} has significantly improved detection accuracy by processing data in an end-to-end fashion. Despite these performance gains, prevailing deep learning methods typically treat deception detection as an opaque binary classification problem. In high-stakes real-world scenarios such as judicial proceedings and security screenings, providing an isolated predictive label is insufficient. Systems must instead offer a verifiable chain of evidence that connects fundamental audiovisual cues including micro-expression variations, vocal tremors, and gaze shifts to the final decision-making process. In this paper, we leverage MLLMs to generate text-based reasoning chains that provide interpretability for the entire process. 

\begin{figure*}[h]
  \centering
  \includegraphics[width=\linewidth,height=9.5cm,keepaspectratio]{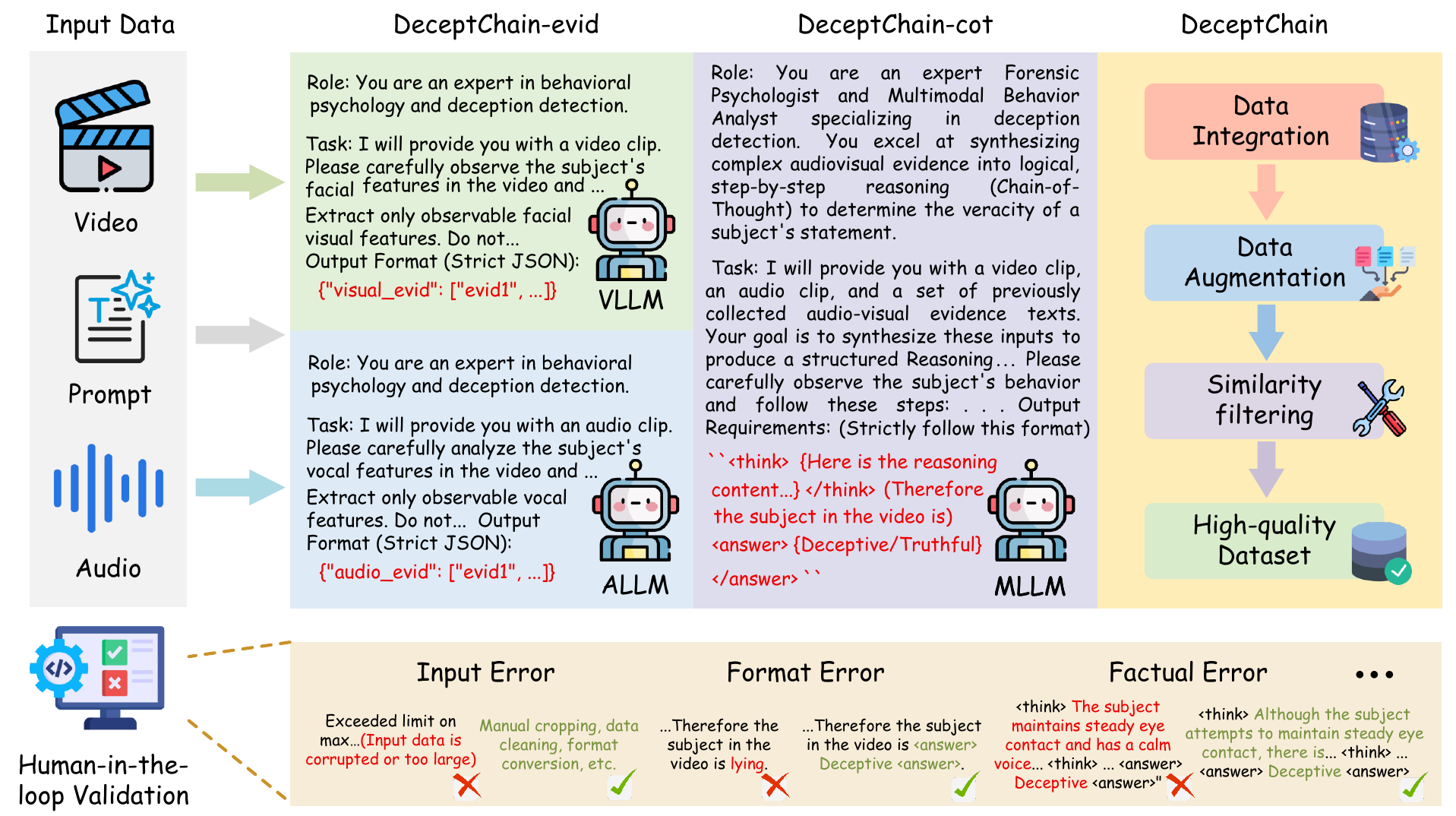}
  \vspace{-0.2cm}
  \caption{The construction pipeline of our proposed instruction datasets~(DeceptChain-evid, DeceptChain-cot, and DeceptChain) follows a model-led and human-assisted annotation strategy designed to ensure high label quality at scale. By leveraging human priors to guide the generation of descriptions and the refinement of samples through a multi-stage pipeline involving LLM-based filtering, keyword heuristics, and manual verification, we achieve the automated annotation of large-scale data.}
  \label{fig:2}
\end{figure*}

\begin{table*}[!t]
    \vspace{-0.2cm}
    \centering
    \caption{Comparison of different multimodal deception detection datasets. ``-'' indicates that there is no relevant information.}
    \label{tab:1}
    \begin{tabular*}{\textwidth}{@{\extracolsep{\fill}}lccccccc@{}}
        \toprule
        \multicolumn{1}{l}{\textbf{Dataset}} & \textbf{Videos} & \textbf{Deceptive} & \textbf{Truthful} & \textbf{Annotation} & \textbf{Setting} & \textbf{Label} & \textbf{Auto } \\ 
        \midrule
        Real Life Trials & 121 & 61 & 60 & - & Courtroom & Binary Classification & \xmark \\
        POLLY & 146 & 73 & 73 & - & Public Speech & Binary Classification & \xmark \\
        Bag of Lies &  325 & 162 & 163 & - &  Laboratory & Binary Classification & \xmark \\
        MU3D & 320 & 160 & 160 & - &  Laboratory & Binary Classification & \xmark \\
        Box of Lies & 1049 &  862 &  187 & - & Game Show & Binary Classification & \xmark \\
        DOLOs &  1675 &  899 & 776 & - & Game Show & Binary Classification & \xmark \\ 
        \midrule
        \textbf{DeceptChain(Ours)} & \textbf{2320} & \textbf{1221} & \textbf{1099} & \textbf{6960} & \textbf{Cross-scene} & \textbf{Reasoning Analysis} & \cmark \\
        \bottomrule
    \end{tabular*}

\end{table*}

\subsection{Multi-modal Large Language Models}

In recent years, Multimodal Large Language Models~(MLLMs)~\cite{cheng2024videollama,chu2023qwen,wang2024qwen2,liu2023visual} have demonstrated exceptional capabilities in cross-modal semantic understanding and high-level reasoning. Unlike traditional deep learning models that are constrained to specific tasks , MLLMs leverage the vast knowledge and linguistic reasoning inherent in large language models~(such as Qwen series~\cite{bai2023qwen,hui2024qwen2}) to process and interpret diverse inputs, including video, audio, and text. Although they have achieved success in general tasks like video captioning and question answering , applying MLLMs to deception detection still faces several unique challenges. First, existing models often produce false or superficial judgments because they lack fine-grained, domain-specific instruction tuning to identify subtle deceptive cues such as micro-expressions or vocal tremors. Second, existing frameworks primarily focus on content understanding rather than the complex, expert-level logical deduction required for behavioral analysis.

To bridge this gap, we propose the DeceptionX framework, which enhances these general MLLM capabilities through a multi-stage pipeline. We also introduce a specialized instruction dataset, DeceptChain, which mimics the human cognitive pattern of ``Observe-Think-Summarize''. This approach directly links low-level audiovisual evidence to the final veracity judgment, ensuring that the model's reasoning is both accurate and verifiable.

\begin{figure*}[h]
  \centering
  \includegraphics[width=\linewidth]{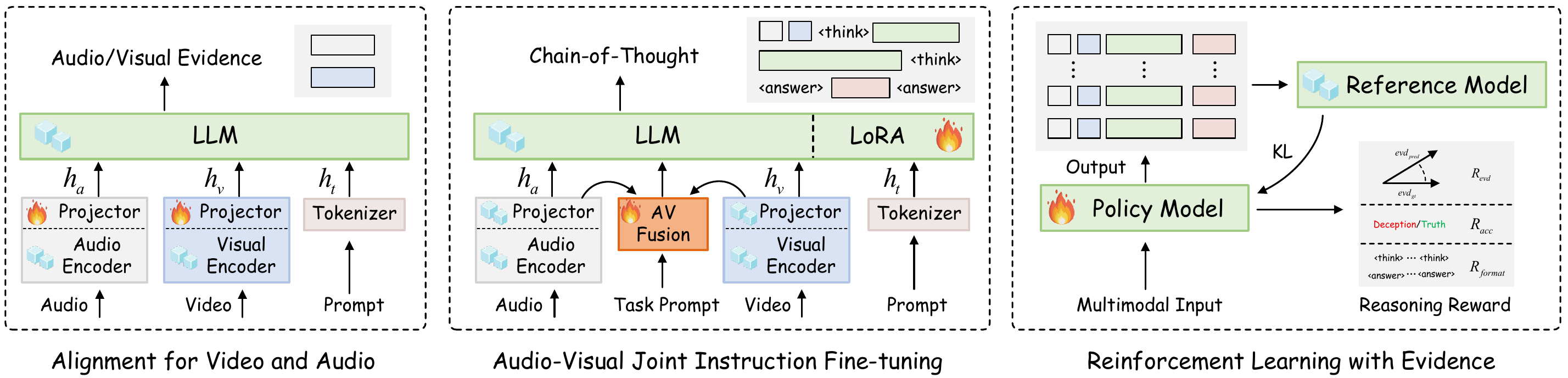}
  \vspace{-0.5cm}
  \caption{The three-stage training framework of DeceptionX: (1) Audio-video feature alignment: align multimodal audio and video evidence into the text feature space. (2) Joint audio-video instruction tuning: instruction-tune the model using chain-of-thought data so it acquires fundamental reasoning abilities about the subject’s behavior. (3) Evidence-based reinforcement learning: optimize the policy model with reasoning-based reward mechanisms to further improve deception detection accuracy and the plausibility of its logical inferences.}
  \label{fig:3}
  \vspace{-0.2cm}
\end{figure*}

\section{DeceptChain}

A significant bottleneck in applying Multimodal Large Language Models (MLLMs) to deception detection is the lack of domain-specific datasets that provide fine-grained reasoning. To bridge this gap, we introduce DeceptChain, a high-quality instruction tuning dataset. As shown in Table~\ref{tab:1}, unlike existing datasets~\cite{soldner2019box,perez2015deception} that restrict deception to a binary classification label, DeceptChain is explicitly designed to support detailed reasoning analysis. It was constructed from existing deceptive-scene video data~\cite{guo2023audio,gupta2019bag,lloyd2019miami} and contains 2,320 videos in total, including 1,221 deceptive samples and 1,099 genuine samples, along with 6,960 rich annotations. To construct this dataset efficiently while ensuring expert-level quality, we design a model-led and human-assisted annotation strategy. As illustrated in Figure~\ref{fig:2}, the data generation pipeline is divided into evidence extraction, chain-of-thought synthesis, and human-in-the-loop validation. More details about the DeceptChain dataset can be found in the supplementary material (Appendix).

\section{DeceptionX}

Our goal is to enhance the deception detection capability of MLLMs. To this end, we adopt the progressive multimodal training paradigm shown in Figure~\ref{fig:3}, designed to enable MLLMs to transition from identifying suspicious behavior to reasoning about and summarizing the deception process. In this section, we introduce DeceptionX, a framework specifically designed for deception detection.

\subsection{Pretraining based on audio-visual evidence}

In the first stage, our objective is not to have the model produce a binary true/false judgment directly, but to first establish a robust multimodal evidence alignment capability that enables it to recognize and extract verifiable low-level cues from raw video and audio. Specifically, the model receives video frames and their corresponding audio, encodes them, and maps them into a unified textual semantic space. During this process, the model is constrained to attend to observable signals that are highly relevant to deceptive behavior, such as microexpressions, gaze shifts, tonal variations, and vocal tremor, thereby forming preliminary correspondences between audiovisual cues and textual semantics. This stage functions as a perceptual foundation for subsequent reasoning, teaching the model to see the evidence before explaining it, and emphasizes strengthening downstream learnability and interpretability through multimodal evidence alignment. Visual and audio signals are processed separately through their respective encoders and projectors, and can be expressed formally as follows:

\begin{equation}
    \begin{aligned}
        h_{a} =\text{Projector}_{a}(\text{Encoder}_{a}(f_a))\\
        h_{v} = \text{Projector}_{v}(\text{Encoder}_{v}(f_v))\\
    \end{aligned}
\end{equation}
The prompt is converted into $h_t$ by the tokenizer. To enhance the alignment of Audio/Video-Language modeling, we freeze LLM and the pre-trained encoder, and fine-tune only the projector.

\begin{figure*}[h]
  \centering
  \includegraphics[width=\linewidth]{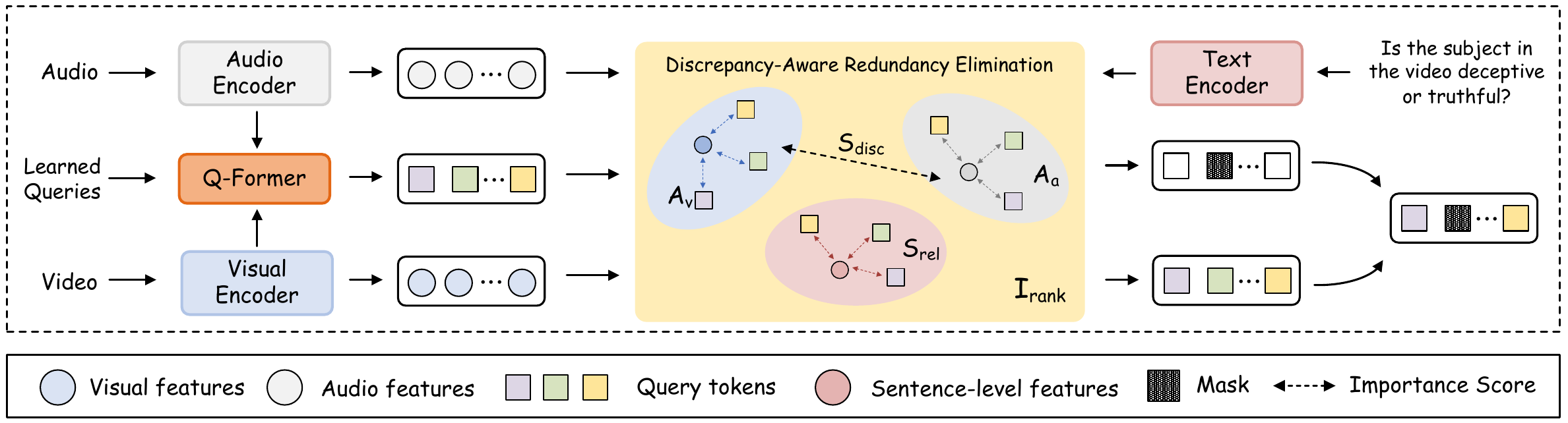}
  \vspace{-0.5cm}
  \caption{Architecture of the AV-fusion and Discrepancy-Aware Redundancy Elimination (DARE). Features are temporally concatenated and fused with learnable queries via a Q-Former to produce compact multimodal representations. DARE then computes sentence-level importance scores to identify suspicious deceptive segments, applies dynamic masking to eliminate redundant or low-salience information (grayed out), and retains only highly relevant cues.}
  \label{fig:4}
  \vspace{-0.3cm}
\end{figure*}

\subsection{Chain-of-thought Supervised Fine-Tuning}

After completing evidence alignment, the second stage introduces Chain-of-Thought (CoT) data for instruction fine-tuning to transition the model from evidence recognition to causal reasoning. Training samples in this stage include not only the raw audio-visual inputs but also the structured evidence texts generated in the previous stage and manually verified. The model is required to analyze suspicious behavior and then explain why these cues jointly indicate deceptive intent using a fixed format such as <think> and <answer>. This training regime does not merely increase output length but explicitly teaches the model to follow an expert reasoning chain of observe, think and summarize, organizing local audiovisual cues into a coherent and verifiable logical process. Additionally, in this stage we introduce an audio-visual fusion module in which temporal information is preserved in the visual features $h_v$ and the audio features $h_a$ and Q-Former~\cite{li2023blip} is used for multimodal fusion. Concretely, to compress the multimodal content, we first create $K$ learnable query tokens $h_q \in R^{K×d}$. Then, through cross-attention, $h_q$ interacts with the concatenated $h_c$ to distill knowledge from the multimodal content into the query tokens. Formally, this process is expressed as:

\begin{equation}
    \begin{aligned}
        h_{c} &= \text{Concat}(h_a,h_v)\\
        h_{f} &= \text{Q-former}(h_q,h_c+\text{PE}(h_c))\\
    \end{aligned}
\end{equation}
where $h_{c} \in \mathbb{R}^{(t_a + t_v) \times d}$, with the concatenation operation 
applied along the temporal dimension. Here, $h_f \in \mathbb{R}^{K \times d}$, and $\text{PE}(\cdot)$ represents the positional encoding.

\subsection{Reinforcement Learning with Evidence}

Although the preceding stages have endowed DeceptionX with robust multimodal evidence alignment and basic chain-of-thought reasoning capabilities, the model still requires explicit optimization to simultaneously maximize detection accuracy and the logical plausibility of its inferences. To this end, we introduce the third training stage~(evidence-based reinforcement learning) that refines the policy model $ \pi_\theta $ (the MLLM after CoT supervised fine-tuning) by directly optimizing a composite reward that evaluates both veracity judgments and their grounding in low-level audiovisual cues. Formally, given a multimodal input $ x = (v, a, t) $~(video frames, audio, and prompt), the policy $ \pi_\theta $ generates structured evidence $ e $ and a reasoning trace $ \langle \text{think} \rangle \cdots \langle \text{think} \rangle $, followed by a final binary label $ y \in \{\text{Deceptive}, \text{Truthful}\} $. Inspired by GRPO~\cite{shao2024deepseekmath}, our reward function consists of three components: a format reward~($\mathcal{R}_{format}$), an accuracy reward~($\mathcal{R}_{acc}$) and an evidence reward~($\mathcal{R}_{evd}$).

\noindent\textbf{Format Reward~($R_{format}$).} To ensure structured outputs, we define a binary reward function that checks whether the model's response contains evidence, a reasoning section, and an answer section. It returns 1.0 if all are present and -1.0 otherwise:

\begin{equation}
\mathcal{R}_{\text{format}}(o|v, a, t) = 
\begin{cases} 
+1.0, & \text{is\_required\_format}(o) \\ 
-1.0, & \text{otherwise} 
\end{cases}
\end{equation}

\noindent\textbf{Accuracy Reward~($R_{acc}$).} To force the DeceptionX to produce a clear detection category, we strictly constrain the output space. The label reward gives a positive reward for exact matches and penalizes incorrect guesses. In addition, if the model fails to output a categorical conclusion, it receives a severe penalty. The formulation is as follows:

\begin{equation}
\mathcal{R}_{\text{acc}}(o|v, a, t) = 
\begin{cases} 
+1.5, & o^{a}=y^{a} \\ 
-1.0, & o^{a}\neq y^{a} \\
-1.5, & o^{a} = \varnothing

\end{cases}
\end{equation}

\noindent\textbf{Evidence Reward~($\mathcal{R}_{evd}$).} To further improve the reasoning accuracy of DeceptionX, we impose constraints on the areas where the model is prone to hallucinations, specifically the audiovisual evidence section. In particular, we define a semantic threshold and reward reasoning samples that exceed this value to establish grounded reasoning behavior. The formulation is as follows:

\begin{equation}
\mathcal{R}_{\text{evd}}(o|v, a, t) = 
\begin{cases} 
\text{Cos}(o^{evd},y^{evd}), & \text{greater\ than\ threshold} \\ 
-1.0, & \text{otherwise}
\end{cases}
\end{equation}
where Cos(·) computes the semantic similarity between the generated evidence and the ground truth. After undergoing three stages of training according to the progressive strategy described above, the MLLM’s deception detection capability is improved.

\begin{table*}[!t]
\centering
\caption{Comparisons of DeceptionX with recent MLLMs. We evaluate performance across three datasets: DOLOs, Bag-of-Lies (BoL), and MU3D.}
\vspace{-0.2cm}
\small
\resizebox{\textwidth}{!}{
    \begin{tabular}{lccccccccccc} 
    \toprule
     & & & \multicolumn{3}{c}{\textbf{DOLOs}} & \multicolumn{3}{c}{\textbf{BoL}} & \multicolumn{3}{c}{\textbf{MU3D}} \\
    \cmidrule(lr){4-6} \cmidrule(lr){7-9} \cmidrule(lr){10-12}
    \textbf{Method/Model} & \textbf{Modalities} & \textbf{Params} & \textbf{Acc.} & \textbf{F1} & \textbf{SEA} & \textbf{Acc.} & \textbf{F1} & \textbf{SEA} & \textbf{Acc.} & \textbf{F1} & \textbf{SEA} \\
    \midrule
    \rowcolor{gray!15} \multicolumn{12}{l}{\textit{Closed Model}} \\
    Qwen-VL-Max~\cite{wang2024qwen2} & Video  & - & 54.32 & 51.38 & 43.42 & 41.78 & 43.76 & 45.52 & 46.28 & 42.52 & 50.86 \\
    GPT-4o~\cite{openai2024gpt4ocard} & Video\ + \ Audio & - &  59.51 & 54.12 & 62.34 & 50.62 & 56.58 & 61.32 & 50.21 & 52.43 & 51.85 \\
    Gemini3Pro~\cite{googledeepmind2026gemini3pro} & Video\ + \ Audio & - & 62.13 & 59.35 & 70.12 & 53.38 & 57.64 & 55.42 & 51.86 & 55.43 & 56.78 \\
    \midrule
    \rowcolor{gray!15} \multicolumn{12}{l}{\textit{Open Source Model}} \\
    Qwen2.5-VL~\cite{bai2025qwen25vltechnicalreport} & Video & 7B & 46.74 & 53.28 & 42.21 & 43.68 & 42.51 & 23.47 & 50.56 & 52.47 & 44.83 \\
    AffectGPT~\cite{lian2025affectgpt} & Video\ + \ Audio & 7B & 64.08 & 66.52 & 68.84 & 55.43 & 57.21 & 69.32 & 54.98 & 57.64 & 68.53 \\
    VideoLLaMA2~\cite{cheng2024videollama} & Video\ + \ Audio & 7B & 54.24 & 57.68 & 60.21 & 45.85 & 50.27 & 52.23 & 50.78 & 53.64 & 54.58 \\
    PandaGPT~\cite{su2023pandagpt} & Video\ + \ Audio & 7B & 52.14 & 53.56 & 57.78 & 47.82 & 50.23 & 49.86 & 51.57 & 54.21 & 55.43 \\
    Qwen2.5-Omni~\cite{xu2025qwen25omnitechnicalreport}& Video\ + \ Audio & 7B & 51.16 & 52.58 & 57.43 & 44.37 & 44.25 & 51.58 & 53.56 & 55.71 & 60.23 \\
    \midrule
    \textbf{DeceptionX}~(Ours) & \textbf{Video\ + \ Audio} & 7B  & \textbf{70.12} & \textbf{73.26} & \textbf{80.32} & \textbf{61.41} & \textbf{63.87} & \textbf{78.63} & \textbf{60.58} & \textbf{61.83} & \textbf{78.67} \\
    \bottomrule
    \end{tabular}
    \label{tab:2}
}
\vspace{-0.3cm}
\end{table*}

\subsection{Redundancy Elimination Strategy}

Although the Q-Former compresses the temporally concatenated audiovisual features $  h_c  $ into a compact set of $  K  $ fused tokens $  \{z_i\}_{i=1}^K  $, long-duration deceptive videos inevitably contain substantial redundant or background information that dilutes model attention on sparse, diagnostically critical cues such as micro-expressions and vocal tremors. To dynamically prioritize suspicious segments while eliminating low-salience tokens, we introduce the Discrepancy-Aware Redundancy Elimination~(DARE). As illustrated in Figure~\ref{fig:4}, DARE is seamlessly integrated after the Q-Former within the AV-fusion module and operates directly on the fused representations before they are fed to the LLM. DARE computes a per-token importance score $  I_{rank}  $ by combining two complementary signals specifically tailored to deception detection: prompt-anchored semantic relevance and cross-modal discrepancy mining.

\begin{algorithm}[!t]
\caption{Discrepancy-Aware Redundancy Elimination}
\label{alg:dare}
\begin{algorithmic}[1]
\REQUIRE Fused tokens $\{z_i\}_{i=1}^{K}$,
prompt features $t_s$,
visual features $\{h^v_t\}_{t=1}^{T}$,
audio features $\{h^a_t\}_{t=1}^{T}$,
attention weights $\{\alpha_{i,t}\}$,
weight $\omega$, masking number $N_{\mathrm{mask}}$
\ENSURE Selected tokens $\mathcal{Z}_{\mathrm{sel}}$

\FOR{each time step $t$}
    \STATE Compute audio-visual discrepancy:
    \STATE \hspace{0.5cm}
    $D_t = 1-\cos(h^v_t,h^a_t)$
\ENDFOR

\FOR{each token $z_i$}
    \STATE Compute semantic relevance:
    \STATE \hspace{0.5cm}
    $S_{\mathrm{rel}}^{(i)}
    =
    \frac{
    z_i^{\top}t_s
    }{
    \|z_i\|\|t_s\|
}$

    \STATE Compute discrepancy score:
    \STATE \hspace{0.5cm}
    $S_{\mathrm{disc}}^{(i)}
    =
    \sum_{t=1}^{T}\alpha_{i,t}D_t$
\ENDFOR

\STATE Normalize
$\{S_{\mathrm{rel}}^{(i)}\}_{i=1}^{K}$
and
$\{S_{\mathrm{disc}}^{(i)}\}_{i=1}^{K}$
to obtain
$\{\widetilde{S}_{\mathrm{rel}}^{(i)}\}_{i=1}^{K}$
and
$\{\widetilde{S}_{\mathrm{disc}}^{(i)}\}_{i=1}^{K}$

\FOR{each token $z_i$}
    \STATE Compute importance:
    \STATE \hspace{0.5cm}
    $I_i
    =
    \omega\widetilde{S}_{\mathrm{rel}}^{(i)}
    +(1-\omega)\widetilde{S}_{\mathrm{disc}}^{(i)}$
\ENDFOR

\STATE Select the indices of the $N_{\mathrm{mask}}$ tokens
with the lowest importance scores:
\STATE \hspace{0.5cm}
$\mathcal{M}
=
\operatorname{BottomK}
\left(
\{I_i\}_{i=1}^{K},
N_{\mathrm{mask}}
\right)$

\STATE Retain the unmasked high-importance tokens:
\STATE \hspace{0.5cm}
$\mathcal{Z}_{\mathrm{sel}}
\leftarrow
\{z_i \mid i\notin\mathcal{M}\}$

\RETURN $\mathcal{Z}_{\mathrm{sel}}$
\end{algorithmic}
\end{algorithm}

\noindent\textbf{Prompt-Anchored Semantic Relevance Assessment.} To suppress tokens that capture irrelevant background noise unrelated to the deception/honesty judgment task, we first evaluate the semantic alignment of each fused token $  z_i  $ with the task-specific prompt features $ t_{s}$ (derived from the input instruction). We compute the cosine similarity between all fused tokens and the sentence-level prompt representation:
\vspace{-0.1cm}

\begin{equation}
S_{\mathrm{rel}}^{(i)}
=
\frac{
z_i^{\top}t_s
}{
\|z_i\|\|t_s\|
}.
\end{equation}
this mechanism explicitly anchors token importance to the core semantics of the deception detection pipeline, ensuring that only cues relevant to behavioral analysis are retained.

\noindent\textbf{Cross-Modal Discrepancy Mining.} A defining characteristic of deceptive behavior is the incongruence between visual and auditory modalities. To explicitly capture such diagnostic contradictions, we measure the differential affinity between the  visual feature sequences $\{h_{t}^{v}\}_{t=1}^{T}$ and the audio feature sequences $\{h_{t}^{a}\}_{t=1}^{T}$. The specific formulation is as follows:

\begin{equation}
D_t = 1 - \cos(h_{t}^{v}, h_{t}^{a})
\end{equation}
The cross-modal discrepancy score is then defined as:


\begin{table}[t]
\centering
\caption{Comparisons of the proposed DeceptionX with recent methods.}
\vspace{-0.2cm}
\resizebox{0.47\textwidth}{!}{ 
    \begin{tabular}{lc|cccccc}
    \toprule
    \multirow{2}{*}{Model} & \multirow{2}{*}{Reason} & \multicolumn{2}{c}{DOLOs} & \multicolumn{2}{c}{BoL} & \multicolumn{2}{c}{MU3D} \\ 
    \cmidrule(lr){3-4} \cmidrule(lr){5-6} \cmidrule(lr){7-8} 
    & & Acc. & F1 & ACC & F1 & Acc. & F1 \\
    \midrule
    MLP & \text{\sffamily X} & 57.49 & - & 49.90 & - & - & - \\
    LieNet & \text{\sffamily X} &  56.50  &     69.72 & 59.78 &  58.14 & 53.48 & 33.62 \\
    FacialCueNet & \text{\sffamily X} & 60.98 &  68.65 & 56.23 &  63.26 &  57.64 & 59.13\\
    DDABG & \text{\sffamily X} & 55.47 & 62.52 &  56.66 & 55.17 &  55.63 & 51.99\\
    PECL & \text{\sffamily X} & 64.75 &  71.20 & 59.51 &  51.06 &  56.25 & 45.72 \\ 
    STRD & \text{\sffamily X} & 66.87 &  71.13 & - &  - &  - & - \\ 
    \midrule
    DeceptionX & \checkmark & \textbf{70.12} & \textbf{73.26} & \textbf{61.41} & \textbf{63.87}  & \textbf{60.58} & \textbf{61.83} \\ 
    \bottomrule
    \end{tabular}
    \label{tab:3}
}
\vspace{-0.6cm}
\end{table}

\begin{equation}
S_{disc}^{(i)} = \sum_{t=1}^{T} \alpha_{i,t} \cdot D_t
\end{equation}
where $\alpha_{i,t}$ is the attention weight of $z_i$ assigned to time step $t$. A larger $D_t$ indicates lower agreement between the synchronized visual and audio representations at time step $t$, suggesting stronger cross-modal discrepancy. Higher values of $ S_{\text{disc}}^{(i)} $ indicate that the token has successfully encoded conflicting multimodal signals, which are precisely the subtle cues human experts rely on to detect deception. The final importance score for each token is obtained via a weighted linear combination after normalization ($  \tilde{S}  $) across all tokens:

\begin{equation}
I_i = \omega \cdot \tilde{S}_{\text{rel}}^{(i)} + (1 - \omega) \cdot \tilde{S}_{\text{disc}}^{(i)}
\end{equation}
The hyperparameter weight $ \omega \in [0,1] $ balances these two signals. The $N_{mask}$ tokens with the lowest importance scores are masked, ensuring that only high-importance representations are forwarded to the LLM for subsequent processing.


\section{Experiments}

In this section, we comprehensively evaluate the effectiveness of the proposed DeceptionX framework on widely-used deception detection benchmarks, including DOLOs~\cite{guo2023audio}, Bag of Lies (BoL)~\cite{gupta2019bag}, and MU3D~\cite{lloyd2019miami}. We report Accuracy (Acc.) and F1 score to evaluate the model's detection performance, and to rigorously evaluate the reasoning capability of our model, we also introduce the Semantic Evidence Alignment (SEA) metric. Unlike traditional n-gram based metrics (e.g., BLEU~\cite{papineni2002bleu}, ROUGE~\cite{lin2004rouge}), which fail to capture semantic variations in descriptive cues (e.g., ``gaze aversion'' or ``looking away''), SEA leverages a pre-trained sentence encoder to map visual and audio cues into a high-dimensional semantic space. Specifically, we compute the cosine similarity matrix between the generated semantic evidence and the ground truth. We define a successful reasoning as a semantic match exceeding a threshold $\tau$, which allows us to calculate the Precision, Recall, and F1-score for evidence grounding. This ensures that our evaluation reflects the model's actual perceptual understanding rather than mere lexical overlap. More details can be found in the appendix.

\subsection{Main Result}

\noindent\textbf{Comparison with MLLM Baselines.} Table~\ref{tab:2} reports the quantitative comparison of DeceptionX against recent closed-source and open-source MLLMs across the three benchmarks. DeceptionX consistently achieves the highest Accuracy and F1 scores on all datasets. For example, on DOLOs it reaches 70.12\% Acc. and 73.26\% F1, outperforming the strongest closed-source baseline Gemini3Pro and the best open-source model AffectGPT. More importantly, DeceptionX delivers substantially superior Semantic Evidence Alignment (SEA). These gains validate the effectiveness of our DeceptChain dataset, the progressive three-stage training pipeline, and the Discrepancy-Aware Redundancy Elimination (DARE) strategy in enabling high-quality, human-like Observe-Think-Summarize reasoning while maintaining strong classification performance.

\noindent\textbf{Comparison with Specialized Deception Detection Methods.} Table~\ref{tab:3} further compares DeceptionX with recent state-of-the-art specialized deception detection models (e.g., MLP~\cite{gupta2019bag}, LieNet~\cite{karnati2021lienet}, FacialCueNet~\cite{nam2023facialcuenet}, PECL~\cite{guo2023audio}, DDABG~\cite{kang2024deception}, and STRD~\cite{shao2026spatio}). All compared baselines treat deception detection as a black-box binary classification task (denoted X under Reason) and therefore cannot provide interpretable reasoning. DeceptionX not only outperforms these methods across every dataset but also supplies transparent, expert-level reasoning paths. The ability to simultaneously deliver superior accuracy and verifiable chain-of-thought explanations demonstrates that shifting from opaque classification to an interpretable MLLM paradigm closes the critical gap between performance and practical usability in high-stakes deception detection scenarios.

\noindent\textbf{Cross-Domain Generalization.} To further assess the generalization capability of DeceptionX across heterogeneous deception scenarios, we conduct cross-domain experiments following the X\&Y→Z protocol (training on any two datasets and testing on the remaining one), as reported in Table\ref{tab:4}. DeceptionX consistently outperforms the specialized baselines in most cross-domain transfer settings. Concretely, it achieves the highest average accuracy of 57.04\% and F1 score of 63.92\%, surpassing the best baseline (PECL) by approximately 3.33\% in accuracy and 0.93\% in F1. Notably, in the challenging D\&B→M transfer, our model reaches 56.42\% Acc. and 58.38\% F1. These results demonstrate that DeceptionX effectively enhances the model's robustness and practical applicability in cross-scene deception detection.

\begin{table}[t]
\centering
\caption{Cross-dataset results on DOLOs (D), Bag-of-Lies (B), and MU3D (M).
``$X\&Y \rightarrow Z$'' denotes training on X and Y, testing on Z.}
\vspace{-0.2cm}
\label{tab:4}

\setlength{\tabcolsep}{5.2pt}
\renewcommand{\arraystretch}{1.05}

\resizebox{0.47\textwidth}{!}{%
\begin{tabular}{@{}l|cc|cc|cc|cc@{}}
\toprule
\multirow{2}{*}{Method}
& \multicolumn{2}{c|}{M\&B $\to$ D}
& \multicolumn{2}{c|}{D\&M $\to$ B}
& \multicolumn{2}{c|}{D\&B $\to$ M}
& \multicolumn{2}{c}{Average} \\
\cmidrule(lr){2-3} \cmidrule(lr){4-5} \cmidrule(lr){6-7} \cmidrule(lr){8-9}
& ACC & F1 
& ACC & F1 
& ACC & F1 
& ACC & F1  \\
\midrule

LieNet
& 54.40 & 68.23 
& 54.69 & 50.51 
& 51.08 & \textbf{59.75} 
& 53.39 & 59.50  \\

DDABG
& 55.26 & 66.41 
& 53.18 & 65.74 
& 52.61 & 55.78 
& 53.68 & 62.64  \\

PECL
& 54.51 & \textbf{69.55} 
& 51.25 & 66.95 
& 55.38 & 52.46 
& 53.71 & 62.99  \\

\textbf{DeceptionX} & \textbf{58.38} & 64.86 & \textbf{56.32} & \textbf{68.51} & \textbf{56.42} & 58.38 & \textbf{57.04} & \textbf{63.92} \\

\bottomrule
\end{tabular}%
}
\vspace{-0.2cm}
\end{table}

\begin{table}[t]
\centering
\caption{Ablation results across different datasets.}
\vspace{-0.2cm}
\resizebox{0.47\textwidth}{!}{ 
    \begin{tabular}{lc|cccccc}
    \toprule
    \multirow{2}{*}{TPS} & \multirow{2}{*}{DARE}
    & \multicolumn{2}{c}{DOLOs}
    & \multicolumn{2}{c}{BoL}
    & \multicolumn{2}{c}{MU3D} \\ 
    \cmidrule(lr){3-4}
    \cmidrule(lr){5-6}
    \cmidrule(lr){7-8} 
    & & Acc. & F1 & Acc. & F1 & Acc. & F1 \\
    \midrule
    \text{\sffamily X} & \text{\sffamily X}
    & 64.92 & 66.90
    & 56.10 & 57.23
    & 55.36 & 58.21 \\

    \text{\sffamily X} & \checkmark
    & 66.07 & 68.32
    & 57.50 & 57.93
    & 56.32 & 58.23 \\

    \checkmark & \text{\sffamily X}
    & 68.87 & 70.50
    & 60.64 & 61.72
    & 59.35 & 60.97 \\

    \checkmark & \checkmark
    & \textbf{70.12} & \textbf{73.26}
    & \textbf{61.41} & \textbf{63.87}
    & \textbf{60.58} & \textbf{61.83} \\
    \bottomrule
    \end{tabular}
    \label{tab:5}
}
\vspace{-0.6cm}
\end{table}

\subsection{Ablation Study}

To thoroughly investigate the contribution of each component in the proposed DeceptionX framework, we perform ablation studies on the DOLOs, BoL, and MU3D datasets. As shown in Table~\ref{tab:5}, we ablate the three-stage progressive strategy (TPS) and the Discrepancy-Aware Redundancy Elimination (DARE) module. The baseline without TPS or DARE yields the lowest performance. Adding DARE alone brings modest but consistent gains by dynamically prioritizing suspicious deceptive segments and suppressing redundant background information. Incorporating the full TPS significantly boosts results across all datasets by progressively establishing multimodal evidence alignment, CoT reasoning, and reinforcement-learning-based optimization. The complete configuration (TPS + DARE) achieves the best performance on every benchmark, confirming that these two components work synergistically to improve detection performance across all three datasets.


\begin{figure}[!t]
  \centering
  \includegraphics[width=0.95\linewidth]{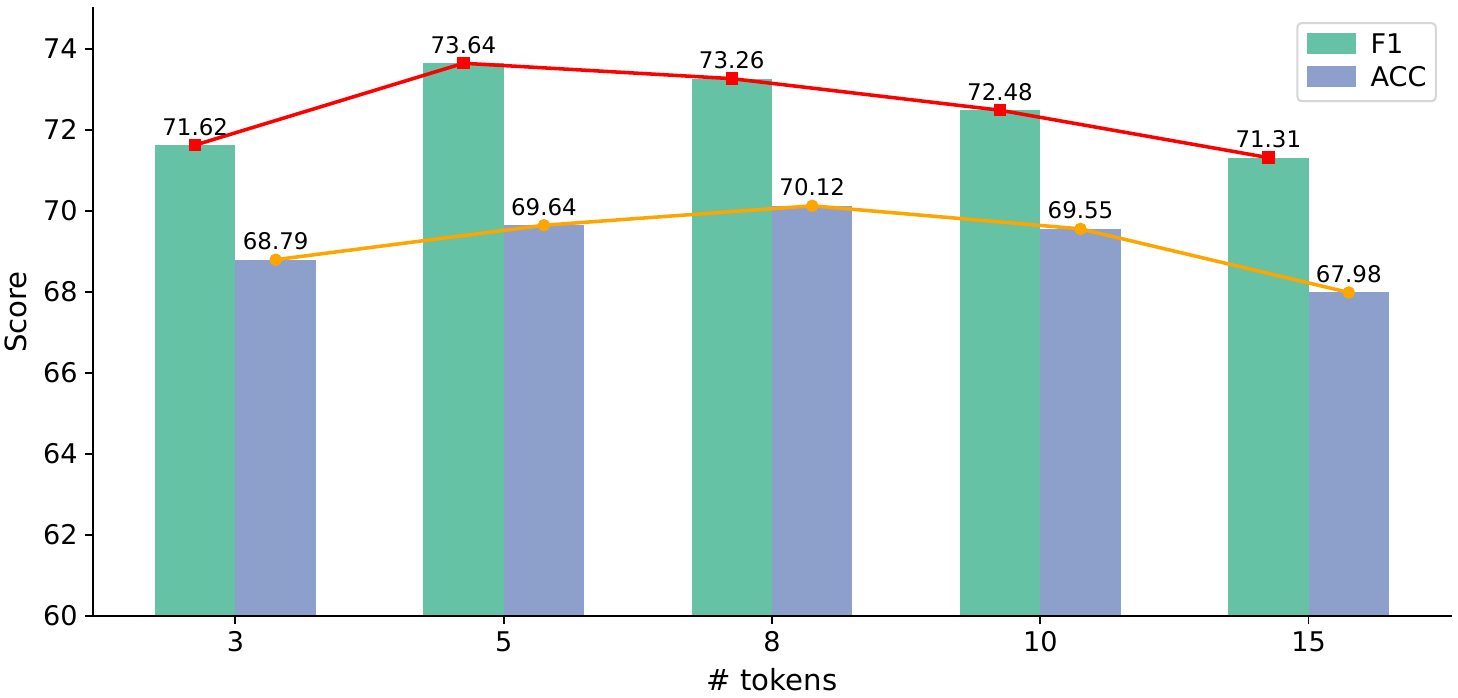}
  \vspace{-0.3cm}
  \caption{Ablation study on masked tokens in DARE.}
  \label{fig:5}
  \vspace{-0.5cm}
\end{figure}

\noindent\textbf{Impact of Masked Token Number in DARE.} To further investigate the behavior of the proposed Discrepancy-Aware Redundancy Elimination (DARE), we analyze the effect of varying the number of masked tokens, as illustrated in Figure~\ref{fig:5}. We vary the number of masked tokens from 3 to 15 and evaluate the performance on DOLOs in terms of Accuracy and F1. The results demonstrate a non-monotonic trend, revealing the trade-off between redundancy removal and information preservation. When the number of masked tokens is small (e.g., 3), the model still retains a large amount of redundant or low-salience information, which distracts the attention from critical deceptive cues and leads to suboptimal performance. As the number of masked tokens increases, the model performance improves and reaches its peak at masking 8 tokens. However, when the number of masked tokens continues to increase, the performance begins to decline. This is because excessive masking removes not only redundant information but also essential discriminative cues. These findings highlight that an appropriate degree of masking is critical for achieving optimal performance.

\section{Conclusion}

In this paper, we propose DeceptionX, a framework that reframes deception detection from a traditional binary classification task into an interpretable reasoning process. To support this framework, we construct a high-quality dataset, DeceptChain, through human-AI collaboration. In addition, we design a progressive training strategy and a redundancy elimination mechanism to enhance the model's perception and reasoning capabilities. Experimental results show that DeceptionX outperforms existing MLLMs and specialized methods across multiple benchmarks, achieving significant improvements in both accuracy and reasoning quality.

\bibliography{aaai2027}


\end{document}